\documentclass[runningheads]{llncs}
\usepackage[T1]{fontenc}
\usepackage{amsmath,amssymb,amsfonts}
\usepackage{algorithmic}
\usepackage{graphicx}
\usepackage{textcomp}
\usepackage{xcolor}
\usepackage{arabtex}
\usepackage{utf8}
\usepackage{tabularx}
\usepackage{array}
\usepackage{ragged2e} 
\usepackage{cite} 
\usepackage{multirow}
\usepackage{hyperref} 
\setcode{utf8}
\hypersetup{
    colorlinks=true, 
    linkcolor=blue,  
    citecolor=blue,  
    urlcolor=blue    
}

\def\BibTeX{{\rm B\kern-.05em{\sc i\kern-.025em b}\kern-.08em
    T\kern-.1667em\lower.7ex\hbox{E}\kern-.125emX}}
    
\begin{document}

\title{A Multi-Layered Large Language Model Framework for Disease Prediction}

\author{
    Malak Mohamed\inst{1}\and
    Rokaia Emad\inst{1}\and
    Ali Hamdi\inst{1} \
}

\institute{
    Faculty of Computer Science, MSA University, Egypt\\
    \email{\{malak.mohamed17, rokaia.emad, ahamdi\}@msa.edu.eg}
}
\authorrunning{M. Mohamed et al} 

\maketitle

\begin{abstract}
Social telehealth has made a breakthrough in
healthcare by allowing patients to share their symptoms and have medical consultations remotely. Users frequently post symptoms on social media and online health platforms, creating a huge repository of medical data that can be leveraged for disease classification and symptom severity assessment. Large language models (LLMs) like LLAMA3, GPT-3.5 Turbo, and BERT process complex medical data, enhancing disease classification. This study explores three Arabic medical text preprocessing techniques: text summarization, text refinement, and Named Entity Recognition (NER). Evaluating CAMeL-BERT, AraBERT, and Asafaya-BERT with LoRA, the best performance was achieved using CAMeL-BERT with NER-augmented text (83\% Type classification, 69\% Severity assessment). Non-fine-tuned models performed poorly (13\%–20\% Type classification, 40\%–49\% Severity assessment). Embedding LLMs in social telehealth enhances diagnostic accuracy and treatment outcomes.
\end{abstract}


\section{Introduction}
The growth of social telehealth has revolutionized the provision of healthcare, enabling patients to share their symptoms and even consult with doctors remotely . During the COVID-19 pandemic, social telehealth increased greatly in popularity; at that time, access to traditional healthcare services was limited. Social media platforms, particularly health forums, are becoming increasingly valuable sources of user-generated medical data that people use to share detailed symptoms and seek the advice of doctors' opinions ~\cite{b1,b2,b4}. However, the unstructured and noisy nature of this data poses a great challenge; hence, advanced computational techniques are required for effective analysis. Recent advances in NLP, especially with the advent of large language models, have really transformed unstructured textual data processing ~\cite{b1,b2,b4}. Various models, including LLAMA, GPT, and BERT, among others, had promised outstanding performance related to text classification and understanding based on large-scale pre-training over big datasets to obtain advanced on a wide range of domains ~\cite{b4,b3,abdellaif2024lmrpa,abdellatif2024lmv}. With these advances, challenges arise in applying LLMs to domain-specific tasks, including healthcare. The fine-tuning of specialized application LLMs have to consider domain-specific nuances, noisy data handling, and optimizing computational efficiency. This is a resource- and computationally intensive process involving new frameworks so that the effectiveness of LLMs in real-world applications is optimized ~\cite{b5,abdellaif2024erpa,hamdi2024riro}.

\begin{figure*}[h]
    \centering
    \includegraphics[width=0.99\textwidth]{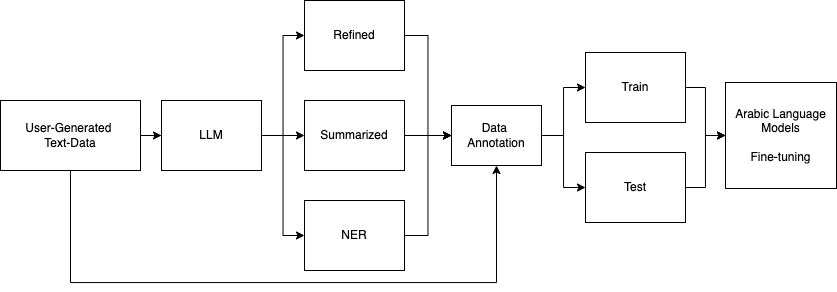}
    \caption{Proposed Multi-Layered Framework for Enhancing Arabic Language Model Fine-Tuning with LLAMA3 Preprocessing}
    \label{fig:image1}
\end{figure*}

We propose a framework that combines LLM-based preprocessing with fine-tuning of Arabic language models for disease classification and severity assessment. It refines text, summarizes posts, and extracts medical entities using NER, enhancing task-specific fine-tuning and overcoming traditional method limitations.

Figure~\ref{fig:image1} illustrates the flow process of our proposed framework. The raw text data generated by users is refined, summarised, NER, and annotated using LLM-based preprocessing. Further, the enriched dataset is utilized to fine-tune Arabic language models for multi-class, multi-label classification tasks, including disease type prediction and symptom severity assessment.

Our contributions in this work are as follows:
\begin{itemize}
    \item \textbf{Novel Integration of LLMs and Arabic Language Models:} We present a multi-layered framework that employs LLMs for preprocessing (refinement, summarization, NER) to enhance the fine-tuning of Arabic language models intended for healthcare applications.
    \item \textbf{Enhanced Preprocessing Pipeline:} We create an improved dataset. LLM-based preprocessing for making challenges related to user-generated content more interpretable and more structured.
    \item \textbf{Improvement of Classification Performance:} We enhanced the accuracies of disease type and severity classification by fine-tuning pre-trained Arabic language models like CAMeL-BERT, AraBERT, and Asafaya-BERT on LLM-preprocessed data.
\end{itemize}

This work emphasizes the transformative potential of exploiting LLMs in order to enhance fine-tuning for domain-specific language models when handling real-world challenges in healthcare.

\section{Related Work}

Recent breakthroughs in text classification and fine-tuning of large language Models have enhanced NLP applications across domains. Transformer-based models such as BERT and its variants have achieved remarkable performance on a variety of tasks such as false information detection, sentiment analysis and radical content classification~\cite{hamdi2024llm,b9,b8,hamad2024asem}. Fine-tuning even on small labeled datasets improves performance and bidirectional context capture. Refinements like RoBERTa's NSP task removal boost domain-specific results, achieving an F1 score of 0.8 in medication detection and 3.929 MAE in eye-tracking tasks ~\cite{b10,b12,b8}.

Efficient light-weight models like DistilBERT, which is 40\% smaller and 60\% faster, retain 97\% BERT's performance, excelling in tasks such as the classification of socio-political news \cite{b11,b8}. ALBERT-Base-v2 optimized memory and speed, achieving a 13.04 exact match score in COVID-19 queries ~\cite{wassim2024llm,b8}. XLM-RoBERTa improved results in processing more than 100 languages using the advances in multilingual processing and outperforming prior models by 23\% in multilingual accuracy ~\cite{b19,b14,b15,b8,b13}. Special adaptations like Electra-Small and BART-Large introduced token substitution and hybrid architectures. Electra-Small excelled in multilingual fake news detection with innovative token substitution strategies ~\cite{b16,b8,hosam2024lmrpa}. Both models achieved high performance in specialized tasks such as medical complaint detection and NER, with BART-Large reducing voice recognition errors by 21.7\% ~\cite{b18,b17,b8}. These advancements highlight LLMs' potential in handling noisy and unstructured text data.

Generative LLMs enhance NLP, excelling in specialized tasks. Fine-tuned models like GPT-3.5 and Mistral-7B surpass baselines by 50\% in F1-macro scores on domain-specific datasets ~\cite{b7}. QLoRA improves memory efficiency ~\cite{b6}, but model variability persists, as seen in GPT-NeoX-20B and Llama2-7B.

Preprocessing techniques such as text refinement, summarization, and Named Entity Recognition (NER), are essential for handling noisy and unstructured text. BERT and RoBERTa have performed extremely well in NER tasks, effectively extracting key entities from complex text~\cite{b10,b8,hamdi2024riro}. Combining preprocessing with fine-tuning enhances classification accuracy, especially when raw data lacks structure or clarity.

Based on this, our study introduces a multi-layered framework that uses LLM preprocessing to improve the fine-tuning of Arabic language models and address noise in data. This enhances the effectiveness of Arabic language models such as CAMeL-BERT, AraBERT, and Asafaya-BERT, in disease classification and severity tasks. It also presents an effective strategy for incorporating LLMs into social telehealth applications.

\section{Dataset Collection}

The dataset used in this study was collected from user-generated posts on an online social platform where patients shared their medical complaints in Arabic. The posts contained detailed information such as the status of chronic diseases, descriptions of symptoms, symptom durations, height, weight, gender, and age, categorized into Type, Severity, and Diagnosis. Structuring and annotation were done under the supervision of a medical advisor to ensure relevance and accuracy.

\begin{figure*}[h]
    \centering
    \begin{minipage}[b]{0.45\textwidth}
        \centering
        \includegraphics[width=\textwidth]{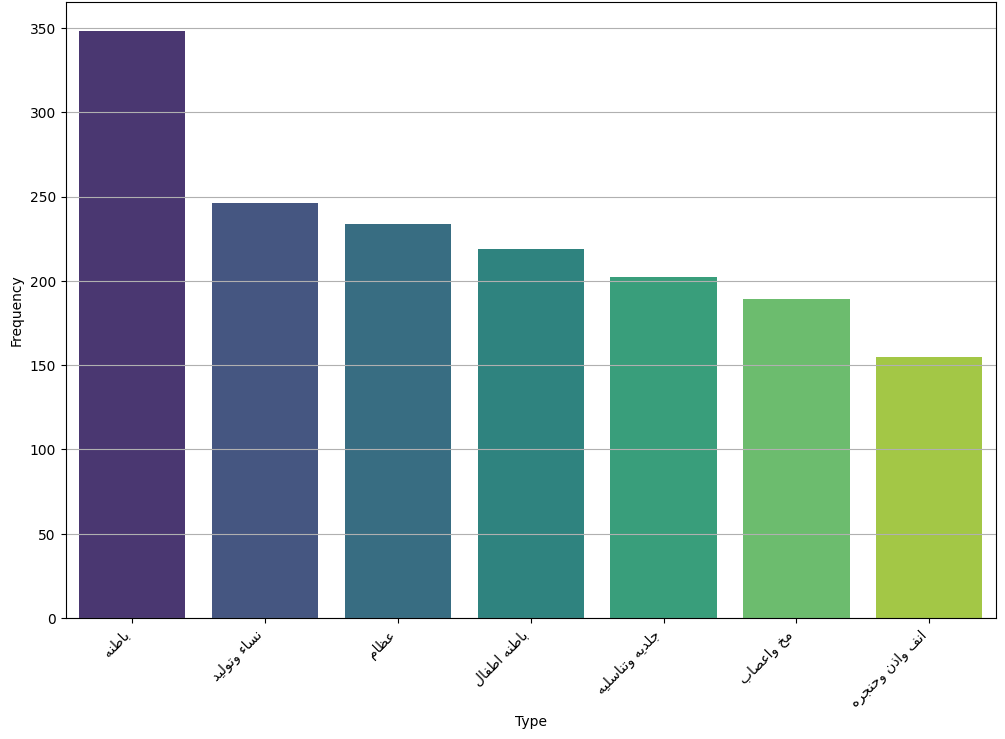}
        \caption{Distribution of condition types in the dataset, illustrating the diversity of medical issues represented.}
        \label{fig:condition_types}
    \end{minipage}
    \hfill
    \begin{minipage}[b]{0.45\textwidth}
        \centering
        \includegraphics[width=\textwidth]{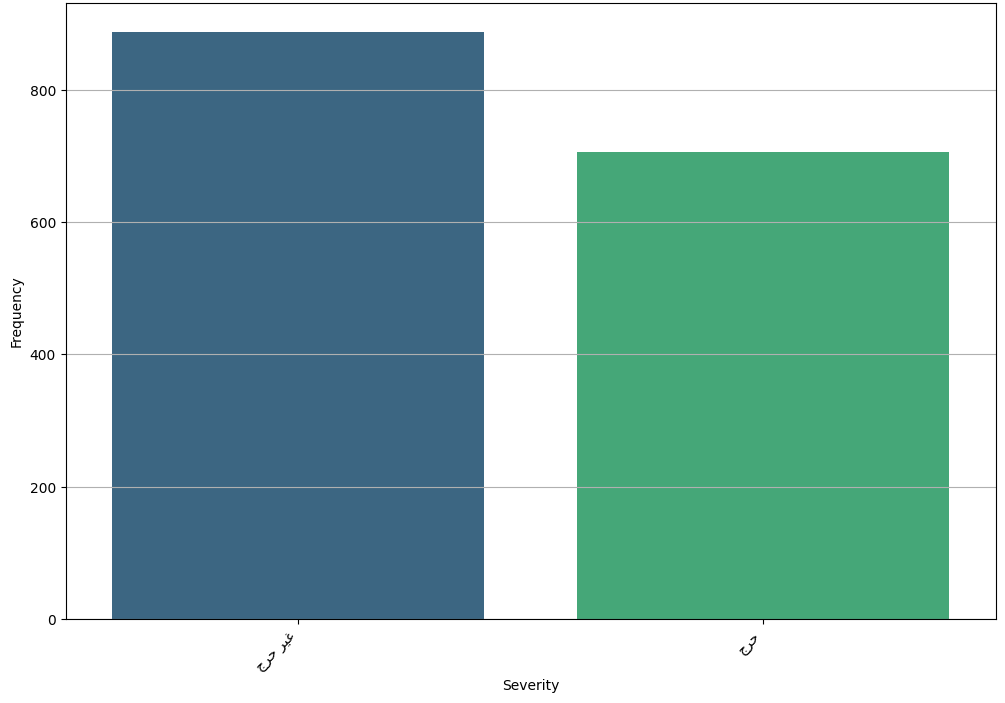}
        \caption{Distribution of severity levels in the dataset, showing the balance between mild and severe cases.}
        \label{fig:severity_levels}
    \end{minipage}
\end{figure*}

Figures \ref{fig:condition_types} and \ref{fig:severity_levels} show the dataset's distribution, with Figure \ref{fig:condition_types} illustrating the variety of medical issues like chronic diseases, skin conditions, and neurological symptoms. Figure \ref{fig:severity_levels} highlights the severity levels, reflecting a balance between mild and severe cases, making these visualizations suitable for a multi-class, multi-label classification task.

\section{Methodology}

This work proposes an Arabic language model fine-tuning using a multi-layered framework. The LLAMA3 model has been used in the enhanced preprocessing step; the flow is illustrated as shown in Fig.~\ref{fig:image1}.

\subsection{Step 1: User-Generated Text Data Collection}
The dataset contains Arabic text data from user-generated content on online health platforms. Texts include elaborate descriptions by the patients of their symptoms, patient medical history, age, and gender. While doing preprocessing, it automatically removes private and sensitive information to ensure privacy.

\subsection{Step 2: Multi-layer preprocessing using LLAMA3}The text is further enhanced using a multi-layered approach to LLAMA 3 filtering. The process involved in preprocessing is as follows:

\begin{itemize}
    \item \textbf{Text Refinement:} LLAMA3 improves the text through deleting unmet requirements such as irrelevant information, grammatical mistakes and the vague parts of the material while protecting the medical context.
    \item \textbf{Text Summarization:}  Llama 3 eliminates unnecessary aspects of long medical posts and compresses them into summaries for easier understanding.
    \item \textbf{Named Entity Recognition (NER):} Symptoms and conditions, together with the drugs, are important medical entities and LLAMA 3 identifies and extracts them.
\end{itemize}

All these steps (refined text, summarized text, and NER-extracted entities) are used to augment the base data.

\subsection{Step 3: Data Annotation}
To create new data variants, the performance data is combined with the outcomes of the preparation stages. These datasets are effective in multi-label classification tasks:
\begin{itemize}
    \item \textbf{Diagnosis Classification:} The goal is to identify the particular medical condition.
    \item \textbf{Type Classification:} This deals with the categorization aspect of the condition (e.g., chronic, acute).
    \item \textbf{Severity Classification:} This category use more qualitative and subjective terminology to assess the the symptoms (e.g., mild, severe).
\end{itemize}

\subsection{Step 4: Arabic Language Models Fine-Tuning}
Three pre-trained Arabic language models are fine-tuned using the improved dataset and they are as follow:
\begin{itemize}
    \item \textbf{CAMeL-Lab/bert-base-arabic-camelbert-mix}
    \item \textbf{aubmindlab/bert-base-arabert}
    \item \textbf{asafaya/bert-base-arabic}
\end{itemize}

The fine-tuning process includes:
\begin{itemize}
    \item \textbf{Optimizing:} Dropout 5\%, scaling by factor 8, rank 16 is some of the settings used along with LoRA.
    \item \textbf{Training Details:} The models have been fine-tuned for a batch size of 4 with 25 epochs. To handle class imbalance, both balanced and accuracy-weighted custom loss functions have been used.
\end{itemize}

\subsection{Step 5: Evaluation}
The fine-tuned Arabic language models are evaluated under
four different preprocessing conditions:
\begin{enumerate}
    \item Normal Text Only
    \item Normal Text + Refined Output
    \item Normal Text + Summarized Output
    \item Normal Text + NER Output
\end{enumerate}

The performance of models is measured concerning accuracy and
balanced accuracy in both disease type classification and
severity prediction.

\begin{table}[!t]
\centering
\caption{Comparison of Text, Refined, Summarized, and NER}
\label{tab:comparison}
\begin{tabular}{|p{6cm}|p{2.5cm}|p{2cm}|p{1.5cm}|}
\hline
\centering\textbf{Text} & \centering\textbf{Refined} & \textbf{Summarized} & 
\textbf{NER} \\\hline
\begin{arabtex}
لو سمحت يا دكتور والدتي عندها 65 سنة مريضة سكر وضغط بتاخد علاج السكر كونكور ٥ تعاني من عدم انتظام في عملية الإخراج مرة إسهال ومرة أخرى إمساك ودلوقتي حاسة أن بطنها منتفخة وميل للقيء أخدت دواء ملين تعمل حمام بعدها وغير كده لا ينتظم معاها، إيه السبب؟ وإيه الدواء المناسب لها؟
\end{arabtex} & 
\begin{arabtex}
العمر 65 سنة الشكوى الطبية عدم انتظام في عمليات الإخراج , اسهال , امساك , حاسة ان بطنها منتفخ وملل للقيء الأعراض سكر ، ضغط ، ميل للقيء
\end{arabtex}
& 
\begin{arabtex}
عدم انتظام في عملية الاخراج مره اسهال ومره امساك حاسه ان بطنها منتفخه وميل للقيء
\end{arabtex}
& 
\begin{arabtex}
عدم انتظام في عمليات الاخراج، اسهال، امساك، حاسة ان بطنها منتفخ و ميل للقيء\end{arabtex} \\ \hline
Excuse me, doctor. My mother is 65 years old, a diabetic and hypertensive patient, and she is taking Concor 5 for her diabetes. She is suffering from irregular bowel movements, sometimes diarrhea and other times constipation. Currently, she feels her abdomen is bloated and has a tendency to vomit. She took a laxative, which helps her use the bathroom, but otherwise, her condition remains irregular. What could be the reason? And what is the appropriate medication for her? 
& Age: 65 years
Medical complaint: Irregular bowel movements (diarrhea, constipation), abdominal bloating, and nausea.
Symptoms: Diabetes, hypertension, bloating, and nausea.
& Irregular bowel movements: sometimes diarrhea, sometimes constipation. She feels her abdomen is bloated and has a tendency to vomit.
& Irregular bowel movements: diarrhea, constipation, feeling of abdominal bloating, and nausea. \\\hline
\end{tabular}
\end{table}

\section{Results and Discussion}

\subsection{Preprocessing with LLAMA3 for Fine-Tuning Enhancement}
Table~\ref{tab:comparison} presents the results of test for Arabic language models that were preprocessed to enhance with LLAMA3. The preprocessing pipeline consisted of text reﬁnement, summarization, and NER, augmenting the original text before ﬁne-tuning. This approach showed that the outputs of LLAMA3 preprocessing integrated into fine-tuned Arabic language models performed noticeably better than models without either fine-tuning or preprocessing.

\subsection{Refined Text Evaluation}
Table~\ref{tab:evaluation-results} for reﬁned text, LLAMA3 preprocessing with Arabic models improved Type classification while
preserving the performance level in Severity classification. In detail, for example, bert-base-arabic-camelbert reached 79\% Type
accuracy and 63\% Severity accuracy (71\% AVG) on normal text.
After refined text preprocessing was applied, Type accuracy increased to 81\%, whereas Severity accuracy remained the same at
61\%, hence a stable average of 71\%.
Non fine-tuned performed
very poorly with Type accuracy between 13\%-19\% and Severity
accuracy ranging between 40\%-49\%. These findings indicate that
that combines LLM preprocessing and fine-tuning to enhance
classification performance.

\begin{table}[!h]
\centering
\caption{Evaluation Results of Models with and without Fine-Tuning for Refined Text.}
\label{tab:evaluation-results}
\begin{tabular}{|p{2cm}|p{3cm}|c|c|c|c|c|c|}
\hline
\multirow{2}{*}{\textbf{Task}} & \multirow{2}{*}{\textbf{Model}} & \multicolumn{3}{c|}{\textbf{Fine-Tuning Acc}} & \multicolumn{3}{c|}{\textbf{W/O Fine-Tuning Acc}} \\ \cline{3-8}
                               &                                         & \textbf{Type} & \textbf{Severity} & \textbf{AVG}   & \textbf{Type}  & \textbf{Severity} & \textbf{AVG} \\ \hline
\multirow{3}{*}{Normal Text}   & Asafaya-bert-base-arabic               & 73\%          & 65\%              & 69\%          & 17\%           & 43\%              & 30\%        \\ \cline{2-8}
                               & Arabertv2                              & 65\%          & 65\%              & 65\%          & 13\%           & 40\%              & 27\%        \\ \cline{2-8}
                               & Bert-base-arabic-camelbert             & 79\%          & 63\%              & 71\%          & 15\%           & 45\%              & 30\%        \\ \hline
\multirow{3}{*}{{With Refined}} & Asafaya-bert-base-arabic               & 76\%          & 66\%              & 71\%          & 19\%           & 40\%              & 30\%        \\ \cline{2-8}
                               & Arabertv2                              & 62\%          & 57\%              & 60\%          & 16\%           & 39\%              & 28\%        \\ \cline{2-8}
                               & Bert-base-arabic-camelbert             & 81\%          & 61\%              & 71\%          & 13\%           & 49\%              & 31\%        \\ \hline
\end{tabular}
\end{table}

\subsection{NER-Enhanced Text Evaluation}
Table~\ref{tab:ner-evaluation-results} underlines the major boost that comes with
NER-enhanced text preprocessing. The best performing fine-
tuned models came with the addition of NER preprocessing.
For instance, bert-base-arabic-camelbert reached 83\% Type
accuracy and 69\% Severity accuracy (76\% AVG),
outperforming all other approaches.
Without fine-tuning, models have seen very limited improve-
ments with NER preprocessing. Type accuracy was in the range
of 15\%-20\% and Severity accuracy was in the range of 40\%-42\%,
which, while
preprocessing yielded better quality data, fine-tuning was necessary to exploit the benefit of improved data.

\begin{table}[!t]
\centering
\caption{Evaluation Results of Models with and without Fine-Tuning for NER Text.}
\label{tab:ner-evaluation-results}
\begin{tabular}{|p{2cm}|p{3cm}|c|c|c|c|c|c|}
\hline
\multirow{2}{*}{\textbf{Task}} & \multirow{2}{*}{\textbf{Model}} & \multicolumn{3}{c|}{\textbf{Fine-Tuning Acc}} & \multicolumn{3}{c|}{\textbf{W/O Fine-Tuning Acc}} \\ \cline{3-8}
                               &                                       & \textbf{Type} & \textbf{Severity} & \textbf{AVG}   & \textbf{Type}  & \textbf{Severity} & \textbf{AVG} \\ \hline
\multirow{3}{*}{Normal Text}   & Asafaya-bert-base-arabic            & 73\%          & 65\%              & 69\%          & 17\%           & 43\%              & 30\%        \\ \cline{2-8}
                               & Arabertv2                           & 65\%          & 65\%              & 65\%          & 13\%           & 40\%              & 27\%        \\ \cline{2-8}
                               & Bert-base-arabic-camelbert          & 79\%          & 63\%              & 71\%          & 15\%           & 45\%              & 30\%        \\ \hline
\multirow{3}{*}{With NER} & Asafaya-bert-base-arabic            & 75\%          & 66\%              & 71\%          & 20\%           & 40\%              & 30\%        \\ \cline{2-8}
                               & Arabertv2                           & 62\%          & 64\%              & 63\%          & 15\%           & 40\%              & 28\%        \\ \cline{2-8}
                               & Bert-base-arabic-camelbert          & 83\%          & 69\%              & 76\%          & 15\%           & 42\%              & 29\%        \\ \hline
\end{tabular}
\end{table}

\subsection{Summarized Text Evaluation}
As seen in Table~\ref{tab:summarized-evaluation-results}, it is clear that summarization gives limited
improvements over the baseline. In bert-base-arabic-
camelbert, the Type accuracy was unchanged at 79\% and the
Severity accuracy increased slightly to 64\% and hence its average is 72\%. This means summarization has much less effect
compared to refinement or NER.
Non-fine-tuned models had little benefit from summariza-
rization, with Type accuracies ranging between 14\%-17\% and Severity
accuracies between 40\%-46\%. These findings suggest that
summarization is of limited utility without fine-tuning.

\begin{table}[!t]
\centering
\caption{Evaluation Results of Models with and without Fine-Tuning for Summarized Text.}
\label{tab:summarized-evaluation-results}
\begin{tabular}{|p{2.5cm}|p{2.5cm}|c|c|c|c|c|c|}
\hline
\multirow{2}{*}{\textbf{Task}} & \multirow{2}{*}{\textbf{Model}} & \multicolumn{3}{c|}{\textbf{Fine-Tuning Acc}} & \multicolumn{3}{c|}{\textbf{W/O Fine-Tuning Acc}} \\ \cline{3-8}
                               &                                            & \textbf{Type} & \textbf{Severity} & \textbf{AVG}   & \textbf{Type}  & \textbf{Severity} & \textbf{AVG} \\ \hline
\multirow{3}{*}{Normal Text}   & Asafaya-bert-base-arabic                  & 73\%          & 65\%              & 69\%          & 17\%           & 43\%              & 30\%        \\ \cline{2-8}
                               & Arabertv2                                 & 65\%          & 65\%              & 65\%          & 13\%           & 40\%              & 27\%        \\ \cline{2-8}
                               & Bert-base-arabic-camelbert                & 79\%          & 63\%              & 71\%          & 15\%           & 45\%              & 30\%        \\ \hline
\multirow{3}{*}{With Summarized} & Asafaya-bert-base-arabic                  & 75\%          & 64\%              & 70\%          & 16\%           & 40\%              & 28\%        \\ \cline{2-8}
                               & Arabertv2                                 & 63\%          & 61\%              & 62\%          & 15\%           & 40\%              & 28\%        \\ \cline{2-8}
                               & Bert-base-arabic-camelbert                & 79\%          & 64\%              & 72\%          & 14\%           & 46\%              & 30\%        \\ \hline
\end{tabular}
\end{table}

\subsection{Comparison Across Approaches}
Across all methods, the NER-enhanced approach yielded the most
notable improvements with 83\% Type accuracy and 69\%
Severity accuracy (76\% AVG) using fine-tuning. This implicates the use of NER as valuable in extracting important
medical information, therefore making the most effective
preprocessing method for augmentation with fine-tuning.

\section{Conclusion}
This work contributes to fine-tuning the Arabic language models for the purpose of disease classification and symptom severity assessment in social telehealth applications. A two-step methodology involving multi-layered preprocessing for text Refinement, Summarization, and Named Entity Recognition (NER) to extract key words of medical posts. The enriched datasets were used to fine-tune three notable Arabic language models: CAMeL-BERT, AraBERT, and Asafaya-BERT, fine-tuned with NER enhanced approaches, outperformed others with 83\% accuracy in disease classification and 69\% in severity estimation. This model improves performance and provides a new benchmark for its application in telehealth. Future studies can extend this framework to other languages.

\section{Acknowledgment}
Heartfelt gratitude is extended to AiTech AU, \textit{AiTech for Artificial Intelligence and Software Development} (\url{https://aitech.net.au}), for funding this research, providing technical support, and enabling its successful completion.

\bibliographystyle{splncs04}
\bibliography{references}

\end{document}